\documentclass{article}

\usepackage{arxiv}

\usepackage[utf8]{inputenc} % allow utf-8 input
\usepackage[T1]{fontenc}    % use 8-bit T1 fonts
\usepackage{hyperref}       % hyperlinks
\usepackage{url}            % simple URL typesetting
\usepackage{booktabs}       % professional-quality tables
\usepackage{amsfonts}       % blackboard math symbols
\usepackage{nicefrac}       % compact symbols for 1/2, etc.
\usepackage{microtype}      % microtypography
\usepackage{lipsum}
\usepackage{xcolor}

\title{The Next Big Thing(s) in Unsupervised Machine Learning: Five Lessons from Infant Learning}

\author{
  Lorijn Zaadnoordijk \\
  Trinity College Institute of Neuroscience\\
  Trinity College Dublin\\
  Dublin, Ireland\\
  \texttt{L.Zaadnoordijk@tcd.ie} \\
  \And
  Tarek R. Besold \\
  Neurocat GmbH \\
  Berlin, Germany \\
  \texttt{tb@neurocat.ai} \\
  \And
  Rhodri Cusack\\
  Trinity College Institute of Neuroscience\\
  Trinity College Dublin\\
  Dublin, Ireland\\
  \texttt{CusackRh@tcd.ie} \\
}

\begin{document}

\maketitle

\begin{abstract}
  After a surge in popularity of supervised Deep Learning, the desire to reduce the dependence on curated, labelled data sets and to leverage the vast quantities of unlabelled data available recently triggered renewed interest in unsupervised learning algorithms. Despite a significantly improved performance due to approaches such as the identification of disentangled latent representations, contrastive learning, and clustering optimisations, the performance of unsupervised machine learning still falls short of its hypothesised potential. Machine learning has previously taken inspiration from neuroscience and cognitive science with great success. However, this has mostly been based on adult learners with access to labels and a vast amount of prior knowledge. In order to push unsupervised machine learning forward, we argue that developmental science of infant cognition might hold the key to unlocking the next generation of unsupervised learning approaches. Conceptually, human infant learning is the closest biological parallel to artificial unsupervised learning, as infants too must learn useful representations from unlabelled data. In contrast to machine learning, these new representations are learned rapidly and from relatively few examples. Moreover, infants learn robust representations that can be used flexibly and efficiently in a number of different tasks and contexts. We identify five crucial factors enabling infants' quality and speed of learning, assess the extent to which these have already been exploited in machine learning, and propose how further adoption of these factors can give rise to previously unseen performance levels in unsupervised learning.
\end{abstract}

\section{Introduction}
\label{sec:intro}

%\begin{quote}
``\textit{The objective of machine learning algorithms is to discover statistical structure in data. In particular, representation-learning algorithms attempt to transform the raw data into a form from which it is easier to perform supervised learning tasks, such as classification. This is particularly important when the classifier receiving this representation as input is linear and when the number of available labeled examples is small.}'' (Mesnil et al., \cite{mesnil2011unsupervised})
%\end{quote}

Early Deep Neural Network (DNN) architectures commonly implemented a paradigm using unsupervised pre-training followed by supervised fine-tuning \cite{bengio2007greedy,hinton2006fast}. %: In a first stage autoencoders were stacked and trained bottom-up in a greedy, layer-wise fashion using an unsupervised learning algorithm, each learning a nonlinear transformation of its input (the output of the previous layer) that captured the main variations in the input. This was followed by a supervised second stage training the top layer and fine-tuning the deep architecture using a supervised training criterion with gradient-based optimisation. 
Inclusion of the unsupervised pre-training stage led to breakthrough improvements in the performance of trained deep models and was the key to unlocking effective training strategies for deep architectures in the mid-2000s (see, e.g., \cite{baldi2012autoencoders,bengio2013representation,erhan2010does} for details). Since the mid-2010s, unsupervised pre-training has become less popular and---especially in computer vision applications---has to a certain degree been superseded by supervised pre-training of models using large-scale labelled data sets (such as ImageNet \cite{russakovsky2015imagenet}). %, again followed by a fine-tuning phase adapting the model to the actual target task. 
Systems implementing this approach have shown impressive performance in a variety of applications (see, e.g., \cite{carreira2017quo,long2015fully,ren2015faster}).
%including object detection \cite{ren2015faster}, image segmentation \cite{long2015fully} and action recognition \cite{carreira2017quo}.
Still, a major drawback of the fully supervised paradigm is its strict dependency on large-scale, well-curated data sets as starting point. %Other than being labor-intensive, it is a challenge to make these data sets representative of the real world. 
Moreover, reasonable doubts have been expressed regarding the efficacy \cite{he2019rethinking} and further expandability \cite{huh2016makes, recht2019imagenet} of the paradigm. These and related considerations over the last few years triggered renewed interest in unsupervised learning within the Deep Learning community (see, e.g., \cite{burgess2018understanding,caron2018deep,chen2020simple}), which inspired our present article.

Unsupervised representation learning attracts considerable interest from neuroscience and cognitive science. DNNs seem to offer a loosely biologically-inspired modelling and analyses tool for the study of the human brain and/or mind, with representation learning playing a central role (see, e.g., \cite{ma2020neural,yamins2016using}).
In fact, there is a longstanding tradition of prominent and active exchanges between machine learning (ML) and these fields, going back at least to Rosenblatt's perceptron \cite{rosenblatt1958perceptron} and its inspiration by ``nerve net'' \cite{mcculloch1943logical}. An increasing number of voices have been calling to again look at animals and humans for insights into how their biological neural machinery works and how these natural-born intelligent systems learn (see, e.g., \cite{hassabis2017neuroscience, lake2017building,ma2020neural,saxe2020if,sinz2019engineering,zador2019critique}. 
%
%Indeed, some of these initiatives already yielded promising results (see, e.g., \cite{anselmi2016unsupervised}).
%
%(see, e.g., work on unsupervised learning of invariant representations modeled after the operations of the human sensory cortex \cite{anselmi2016unsupervised}) once again proving the value of this type of cross-disciplinary exchange of ideas. 
What is common to such contributions is that they are primarily taking mature human cognizers and their brains as a conceptual starting point. In contrast, we advocate to focus on results from the study of infants and their development: Human infants are in many ways a close counterpart to a computational system learning in an unsupervised manner as infants too must learn useful representations from unlabelled data. However, %unlike ML algorithms, 
infants' representations are learned rapidly, with relatively few examples, and can be used flexibly and efficiently in various different tasks and contexts. Developmental science, the field studying how cognitive capacities develop in human infants and children, has identified several facilitating factors %and mechanisms 
for this surprising performance, which merit discussion also in the ML context. Specifically, they may hold the answers to some of the long-standing %, topical 
questions of representation learning (see, e.g., \cite{bengio2012deep}): What can a good representation buy us? What is a good representation in the first place? And what training principles can help discovering such good representations? 

We are not the first %or the only 
ones to make recourse to developmental research in an ML context (see, e.g., \cite{kidd2019how,tenenbaum2011grow}). However, these prior efforts remain on a very general level regarding their engagement with the relevant insights from developmental science. In contrast, we give specific suggestions for which aspects of infant learning ML researchers should consider and discuss to what extent current ML research is already---consciously or by coincidence---working towards integrating these insights. In doing so, we will mostly limit our focus to developmental learning during the first 12 months of life. This is the period in an infants’ life that would be most similar to what is happening during unsupervised ML. In Sec.~\ref{sec:lessons}, we outline core results from five domains within current developmental science that have been crucial for understanding how infants learn and that offer valuable inspiration for the future of unsupervised ML. Sec.~\ref{sec:discussion} then provides a concluding general discussion of relevant chances and challenges arising in the interaction between ML as (mostly) ``computationally-minded''  field and developmental science as (predominantly) ``psychologically-minded'' discipline. 

\section{Five lessons from developmental science: Key enablers of growing a mind}
\label{sec:lessons}
We discuss five main lessons synthesised from the current state of knowledge regarding infant learning and discuss the direct relevance of these insights for the development of the next generation of unsupervised approaches in ML. Sec.~\ref{subsec:architecture} zooms in on the starting conditions for unsupervised learning in infants, for instance tying into recent work on pre-wiring and pre-training of DNNs. Sec.~\ref{subsec:multiplestreams} summarises insights about cross-modal statistical learning in infants and establishes the connection to corresponding efforts in ML. Sec.~\ref{subsec:structuredinput} provides an overview of developmental research into the temporal structure of infant learning %(which turns out to be far from arbitrary) 
and connects directly to results such as non-stationarity in continual representation learning. Sec.~\ref{subsec:activelearning} discusses results from the study of active learning in infants and its ML parallel. Finally, Sec.~\ref{subsec:sociallearning} addresses the role other agents play in infant learning and what this implies for the computational counterpart thereof.

\subsection{Babies' information processing is guided and constrained from birth}
\label{subsec:architecture}

It is a common---but mistaken---belief that infants’ brain architecture is highly immature. Although the infant brain still grows and develops, a lot of its ultimate structure is already present very early on. For instance, neuroimaging data show that the structural connectivity patterns observed in early infancy are very similar to adult structural connectivity \cite{cabral2019category}. Regarding functional connectivity adult-like patterns can also be observed in infants \cite{doria2010emergence, kamps2020connectivity} even for networks that support cognitive capacities that are not yet behaviourally manifested such as speech \cite{cusack2018does}. Functional visual cortex activation of abstract categories (faces and scenes) is adult-like in its spatial organisation in infants as young as four months of age and is refined in its response pattern through development \cite{deen2017organization}. Moreover, functional connectivity patterns in neonates have been shown to be predictive of later development (e.g., \cite{linke2018disruption}). Compared to adults, infants' neural structure is still more plastic and can change dramatically depending on the type of input it receives; for instance, research of congenitally blind individuals shows that the visual cortex can be re-purposed for other, non-visual, capacities but that this does not happen in those that become blind later in life \cite{bedny2017evidence}. Even so, infants' brain structure is already fairly determined as a whole, thereby constraining processing and guiding learning.

\cite{zador2019critique} argued that creators of neural networks should take into consideration that the highly structured neural connectivity is what allows animals to learn so rapidly. Indeed, ML researchers have been exploring the impact of pre-wiring on network learning. A recent example is work in the context of learning general identity rules. Answering a challenge posed in \cite{marcus1999rule} regarding potential limitations on the generalisation capabilities of recurrent neural networks, \cite{alhama2018pre} used ``delay lines” (a concept first suggested in neuroscience \cite{jeffress1948place}) in an Echo State Network \cite{jaeger2001echo}, i.e., organising a sub-network of neurons in the network in layers, with connections with weights 1 between corresponding nodes and 0 everywhere else connecting each layer to the next. The authors show that intentionally introducing delay lines constituted a significant step towards addressing the hypothesised limitations. The findings from developmental science outlined in the previous paragraph suggest that infant neuroscience offers a rich variety of further inspirations for architectural building blocks of neural~networks.

Turning from architecture to `pre-programmed' capabilities, the developmental science community is still heavily debating which cognitive capacities might already be present at birth and what has to be learned through experience (see e.g., \cite{smith1999infants, spelke1995initial}). As such, many appeals to inborn knowledge as explanation for empirical observations are heavily contested. However, there are some findings of biases in newborns that are generally agreed upon (even if they are interpreted differently across labs). Biases in this regard differ from knowledge in that they are an integral part of the learning mechanism rather than an input to the learning algorithm. What initially was considered a bias for faces or face-like stimuli (consisting of 3 dots in an upside-down triangle configuration) has been shown to be a general attentional bias for top-heavy visual information \cite{simion2011processing,viola2004can}, which develops into a preference for faces within the first months \cite{chien2011no, ichikawa2013two}. Newborns further seem to have a preference for some aspect of biological motion \cite{simion2011processing}, though the exact bias mechanisms have yet to be explained. Also in non-visual domains, early processing biases are found. Newborns discriminate \cite{pena2003sounds} and prefer \cite{vouloumanos2007listening} speech versus non-speech, and prefer infant-directed speech over regular speech \cite{cooper1990preference}. However, these speech biases may in part have been shaped by auditory experiences \textit{in utero}.

Going beyond the general notion of inductive bias in ML architectures \cite{michalski1983theory,mitchell1980need}, biases already play an important role in the training of DNN architectures. Already in the early days of Deep Learning, the initial unsupervised pre-training stage---in a way similar to regularisation---also served the purpose of introducing a particular type of `starting bias' into the architecture, namely a reduction in variance and a shift in the parameter space towards configurations that are useful for supervised learning \cite{erhan2010does}. %\footnote{This places unsupervised pre-training within the family of semi-supervised methods in that it defines a particular initialization point for standard supervised training.}
Still, recently the study of the potential development and influence of particular biases in networks has enjoyed increasing interest, for example regarding the role of shape bias within networks in performing visual tasks \cite{feinman2018learning} or in the context of learning identity relationships with different network architectures \cite{kopparti2020weight}. In the latter work, the authors show that Relation Based Patterns \cite{weyde2019modelling} can be implemented as a Bayesian prior on the network weights, helping to overcome limitations neural networks frequently exhibit related to the learning of identity rules and to generalisation beyond the training data set. Looking at examples from computer vision, the striking performance of a model that was trained to be sensitive to mover events (i.e., events of a moving image region causing a stationary region to move or change after contact)---a parallel to infants' bias for biological movement---in recognising hands and gaze direction from video input demonstrated the usefulness of combining learning and innate mechanisms \cite{ullman2012simple}. Given these early successes and the insights from developmental science, we believe that there is overwhelming evidence for the role biases can play in guiding and augmenting the training of neural networks. 

\textbf{TL;DR:} Even when not invoking rich interpretations of newborn cognition, the earliest processing of information after birth is constrained by the neural architecture and guided by cognitive predispositions. Translating these insights from developmental science to the ML world, they find a direct counterpart in the importance of starting conditions and the growing efforts invested in the study of pre-wiring and pre-training of neural networks. Developmental science suggests that not only the type of ML architecture selected and corresponding network model, the training algorithm or the training data matter, but that the particular instantiation of the architecture and setup of the network play an important role as enabler of efficient and effective training. Not all inductive biases have been made equal. Infants do not have to start their developmental trajectory from a complete \textit{tabula rasa} or from an arbitrary starting setup---and neither should neural networks. 

\subsection{Babies are learning statistical relations across diverse inputs}
\label{subsec:multiplestreams}

There is an abundance of literature studying babies' capacity for detecting and learning from statistical regularities in their physical and social environment. Infants readily learn associations, such as sensorimotor contingencies \cite{jacquey2019detection}, incorrect handling of objects \cite{hunnius2010early}, and associations between voices and faces \cite{brookes2001three}. Interest in more complex statistical learning emerged initially in the language domain when researchers showed that infants can use statistical properties in the language input to detect a wide range of phenomena such as word segmentation structures \cite{saffran1996statistical}, phoneme distributions \cite{maye2002infant}, or non-adjacent dependencies \cite{gomez2005developmental, marcus1999rule}. Regarding domains other than language, infants pick up statistical regularities in visual stimuli \cite{kirkham2002visual} and in action sequences \cite{monroy2019sensitivity}. Probabilistic relations have also been shown to affect infants' attention \cite{tummeltshammer2013learning}. 
%Bayesian theories of learning \cite{lake2017building, tenenbaum2011grow} and Predictive Processing theories \cite{clark2013whatever, friston2010free,koster2020making} have recently been suggested as underlying mechanisms. 
Infants' predictions depend on the distribution in a sample as well as the sampling method \cite{gweon2010infants}, and infants are sensitive to the difference in likelihood between events \cite{kayhan2018infants}. Moreover, while 9-month-olds update their model regardless of the informational value of a cue \cite{kayhan2019nine}, 14-month-old infants represent the statistics of their environment and use new cues to update their model as a function of its informational value \cite{kayhan2019infants}. Indeed, the neural dynamics of infants' prediction and surprise change as a function of priors based on previous experience \cite{kouider2015neural}. Combined, these studies demonstrate that statistical learning in infancy happens across different domains. Furthermore, electrophysiological evidence suggests that this type of learning mechanism is already active in newborns \cite{teinonen2009statistical}.

%The ability to pick up statistical regularities does not in itself differentiate infants from ML algorithms, but infants have many sources of input that are currently largely absent in common ML training routines.
A major difference between statistical learning in infants and in machines resides in the multi- and cross-modal nature of infants' learning processes. In everyday life, infants encounter many situations in which they have to learn from and integrate signals from multiple modalities. Many studies have shown that infants are sensitive to the statistical relations between, for example, visual and auditory (linguistic and non-linguistic) stimuli \cite{bristow2008hearing, brookes2001three, spelke1979perceiving}, visual and tactile stimuli \cite{bremner2008spatial, zmyj2011detection}, and auditory and tactile stimuli \cite{tanaka2018integration}. These multimodal associations are thought to develop via a combination of brain maturation and multimodal experience, which allow for the detection of temporal synchrony (see e.g., \cite{lewkowicz2000development}). The importance of experience was shown in individuals who were temporarily deaf before they received a cochlear implant: They showed decreased audiotactile \cite{landry2013temporary} and audiovisual \cite{stevenson2017multisensory} integration, even after their hearing had been restored. Beyond standard multimodal integration, infants have been shown to be sensitive to the relation between exteroceptive (e.g. visual stimuli) and interoceptive (bodily) signals, such as heart rate \cite{maister2017neurobehavioral}. 
%Infants have further shown to react with increased physiological arousal to snakes and spiders, compared to flowers and fish \cite{hoehl2017itsy}, which has been proposed to be a consequence of both evolutionary predispositions as well as correlations between these threat stimuli and fearful facial reactions of other agents \cite{hoehl2017infants}. 
Furthermore, infants react physiologically to threat stimuli \cite{hoehl2017itsy} and are able to associate these stimuli to fearful facial reactions of other agents \cite{hoehl2017infants}. Finally, multiple infant studies demonstrated that arousal and excitement cause infants to engage with and learn from their environment differently and more extensively (see e.g., \cite{rochat2000perceived}). Physiological and emotional signals thus provide infants with additional learning opportunities.

Infants' processing of signals from diverse inputs likely leads to improved representations and task performance. Multi-modal information can support the disambiguation of conflicting or seemingly incoherent input otherwise obtained from a single sensory stream (e.g. \cite{weatherhead2017read}). It further enables the performance of tasks for which a single type of input is not sufficient. However, importantly, even representations that seem related to one sensory domain benefit from input from other modalities. \cite{cappagli2017auditory} have shown that spatial representations (e.g., distance) for both auditory as well as proprioceptive stimuli are impaired for congenitally blind children and adults, suggesting that visual input is important for these non-visual representations. This work hints at a much greater need for learning from diverse multimodal inputs than one might intuitively consider necessary for unimodal tasks.

ML researchers are already actively exploiting the advantages multi- or crossmodal information processing offers. Multimodal processing has been exploited, for instance, in robotics in object categorisation tasks \cite{marton2011combined,nakamura2007multimodal}, and at the intersection between the vision and the language domain \cite{mogadala2019trends}, including in emotion recognition systems \cite{barbieri2019towards,tzirakis2017end} and in movie summarisation tasks \cite{evangelopoulos2013multimodal}. Regarding unsupervised learning, early successes have been achieved in the late 1990s, for example by performing category learning through multimodal sensing \cite{desa1998category}. Following the early successes of Deep Learning, multimodality also moved into the focus of some researchers in the DNN community, again spanning a wide range of application scenarios from image synthesis \cite{radford2016unsupervised} to unsupervised robot perception \cite{droniou2015deep} and, very recently, image captioning \cite{feng2019unsupervised}. However, whilst the advantages of cross-modal feature learning had already been identified almost a decade ago \cite{ngiam2011multimodal,srivastava2012multimodal}, most current DNNs are still being trained on unimodal data. The recent surge in interest in contrastive learning (see, e.g., \cite{bachman2019learning,chen2020simple}) suggests that this might be about to change. It is worth noting that state-of-the-art contributions such as \cite{tian2019contrastive} make explicit reference to the structure and performance of human multimodal information processing as inspiration and motivation for the approach.

\textbf{TL;DR:} Infants learn statistical relations across diverse multimodal input streams and the resulting representations provably benefit from these richer sources of information. Multimodal approaches have also successfully been pursued in ML already for decades. However, until today these successes have not caused a widespread shift from unimodal to multimodal training of DNNs. The recent surge in interest in contrastive learning in a multiview setting might finally trigger wider adoption of multimodal representation learning more generally, even for unimodal tasks.

\subsection{Babies' input is scaffolded in time}
\label{subsec:structuredinput}

``\textit{Each new sensorimotor achievement---rolling over, reaching, crawling, walking, manipulating objects---opens and closes gates, selecting from the external environment different datasets to enter the internal system for learning.}'' (Smith, \cite{smith2018developing}, p. 326)
Babies are not only learning but, importantly, are also developing (i.e. changing over time). The type of input infants receive is critically dependent on what they can do at any given point in time. A newborn infant will primarily see whatever their caregivers bring into their visual field, a crawling infant will get extensive visual input of the floor, and a sitting infant will be able to see as far into the distance as a walking infant, but does not get the experience of optical flow whilst seated. This also changes the level and possibilities of exploration of the environment when infants transition from one motor ability to the next. For example, walkers can access distant objects, carry these objects and are more likely to approach their mothers to share the objects, while crawlers are more likely to remain stationary and explore objects close to them \cite{karasik2011transition}. Furthermore, differences in posture (e.g. sitting vs lying in supine or prone position) affect infants' possibilities for object exploration \cite{lobo2014not, soska2014postural}. Sensory input goes through developmental changes unrelated to motor development too. For example, newborns' visual acuity is low and gradually increases in the first six months of life \cite{dobson1978visual, sokol1978measurement}. These changes in sensory and motor possibilities do not only allow infants to explore different aspects of their environment, they also drive an expansion of the range of obtained inputs in the direction of increasingly varied stimuli occurring in increasingly complex combinations, thereby introducing a phased-structure in infants' learning experience.

Exactly how much the input is constrained has become more clear through recent experimentation with head-mounted cameras (see e.g., \cite{fiser2006infants, franchak2011head, yoshida2008s}). For example, \cite{fausey2016faces} showed that in approximately a quarter of 1- to 24-month-old infants’ visual input there is another person present (as measured by the presence of at least a face or a hand). Importantly, the frequency of faces vs hands significantly changed as a function of age: the proportion of faces in view declined with age and the proportion of hands in view increased with age. In another study, these authors found that in the first year of life there were relatively few different faces in the infant's view, that they were generally closeby and thus visually large, and that mostly both eyes were visible \cite{jayaraman2015faces}. Providing a conceptual counterpart to these findings in computer vision, improved performance and generalisation of DNNs with low initial visual acuity---obtained by starting the network training with blurred rather than with high-resolution images---corroborate the idea that phased sensory input indeed improves learning \cite{vogelsang2018potential}. 

The general idea of applying a structured learning curriculum---as naturally experienced by infants---to ML systems had already been put forward by \cite{elman1993learning} and in the Deep Learning literature was prominently addressed under the headline of curriculum learning \cite{bengio2009curriculum}. One of the main challenges in the context of providing a structured training scheme for neural networks is the high sensitivity of the curriculum's effectiveness to the mode of progression through the learning tasks, i.e., the syllabus. %(We will return to this problem in Sec.~\ref{subsec:activelearning}.) 
We zoom in on two of the factors influencing this sensitivity, namely critical learning periods and catastrophic interference. Critical periods, which are well-documented in biological learners, reflect a moment of peak plasticity during a specific developmental state (often early in life) that are followed by reduced plasticity (see e.g. \cite{newport2001critical} for review and reflection). Critical periods can be contrasted to open-ended learning systems or systems in which plasticity increases with maturation or experience \cite{newport2001critical}. 
\cite{achille2019critical} show that DNNs also exhibit critical periods during which a temporary stimulus deficit can impair the future performance of the network even to an unrecoverable degree. Strong connections that are optimal relative
to the input data distribution are created during the initial epochs of training (i.e., a ``memorisation phase'') and appear to remain relatively unchanged during additional training. As such, the initial learning transient plays a key role in determining the outcome of the training process, and shortcomings or biases, for instance in the variety of input samples, during early training may not be recovered during the remainder of the training process.

As a consequence of the stability/plasticity dilemma \cite{carpenter1988art}, neural networks can suffer from catastrophic interference; a process where new knowledge overwrites rather than integrates previous knowledge \cite{french1999catastrophic}. Catastrophic interference does not only make it challenging to learn tasks sequentially while maintaining performance, it also has consequences for the order of training stimuli. If a neural network is first trained on all exemplars of one class, and then on all exemplars of another class, it often will not properly retain knowledge about the first class. Different solutions to these problems have been proposed over time, including rehearsal and pseudo-rehearsal learning \cite{robins1995catastrophic}, the use of pairs of plastic and elastic network weights \cite{hinton1987using}, and brain-inspired approaches suggesting the use of dual-memory architectures \cite{kemker2017fearnet} or building on synaptic consolidation \cite{kirkpatrick2017overcoming}. Regarding unsupervised learning in particular, catastrophic interference has, among others, been addressed in the context of continuous and lifelong learning, including the use of undercomplete autoencoders trained for feature transfer across tasks \cite{rannen2017encoder} or approaches motivated by results from neuroscience such as neurogenesis deep learning \cite{draelos2017neurogenesis}. Relatedly, the explicit (meta-)learning of representations for continual learning that avoid catastrophic interference has been proposed \cite{javed2019meta}. Still, the general problem of catastrophic interference remains unsolved~\cite{kemker2018measuring}.

\textbf{TL;DR:} Infants' development leads to a phased-structure in their learning input. This creates a naturally guided curriculum for infant learning. Recognising the potential of this type of learning, ML researchers have attempted to integrate phased inputs in their training regimes. However, thus far, these initiatives are hampered by problems such as catastrophic interference. The quest to achieve continual learning without overwriting previously acquired knowledge thus remains unfinished.

\subsection{Babies actively seek out learning opportunities}
\label{subsec:activelearning}

Infant learning does not just happen passively. Infants play an active role in directing their attention to stimuli from which they learn. This process has been given different names including curiosity-driven learning, active learning, and learning by intrinsic motivation. Curiosity is taken to be a state of arousal that requires actions to modulate the aroused state \cite{berlyne1954experimental}, with the degree of novelty determining infants’ ability to learn \cite{berlyne1960conflict}. The spectrum of arousal is subdivided into the three zones relaxation (i.e., insufficient arousal), curiosity (i.e., optimal for learning), and anxiety (i.e., too much arousal) \cite{day1982curiosity}, where relaxation and anxiety are considered to create little opportunity for learning. Empirical research shows that infants attend significantly longer to stimuli that are at an intermediate level of complexity; a finding that was dubbed ‘the Goldilocks effect’ \cite{kidd2012goldilocks, kidd2014goldilocks}. Related to the U-shaped curve of attention as a function of novelty or complexity, infants vary in their familiarity or novelty preferences to stimuli. This is thought to be dependent on the degree of encoding of the stimuli \cite{hunter1988multifactor}. If the encoding is not yet complete, infants will show a familiarity preference, and move to a novelty preference once encoding is completed. These ideas have led to a plethora of looking time studies. While the richness of the interpretation of looking time studies can be questioned (e.g. \cite{aslin2007s, haith1998put}), the theories on active learning and methods to investigate stimulus encoding have provided important insights into the nature of infants' learning mechanisms.

Active learning also has a long history in ML (see, e.g., \cite{settles2009active,settles2011theories}). ML researchers recognised that an algorithm may learn better and more efficiently if it is allowed to select the data from which it learns \cite{settles2009active}. Over the years, different types of curiosity mechanisms have been proposed for artificial systems. Some researchers suggest that curiosity could be prediction-based, causing agents to attend to input for which predictability is minimal \cite{botvinick2009hierarchically} or maximal \cite{lefort2015active}. In the context of curriculum learning (also see Sec.~\ref{subsec:structuredinput}), \cite{graves2017automated} proposed a multi-armed bandit-based approach to finding progress-maximising stochastic policies over different learning tasks. More closely related to the findings in developmental science, \cite{schmidhuber2008driven} argued that curiosity-driven learning occurs most optimally when the agent seeks out information as a function of its compressibility. %An agent would learn best from generating action sequences that yield previously unknown but quickly learnable information. 
Furthermore, it has been suggested that active learning is driven by a goal to maximise learning progress by interacting with the environment in a novel manner \cite{oudeyer2018computational, oudeyer2007intrinsic}. Using autoencoder networks, computational modelling approaches that compared presenting stimuli in a fixed order or allowing the model to choose its own input showed that maximal learning happens when the model can maximise stimulus novelty relative to its internal states \cite{twomey2018curiosity}. This work emphasised the importance of the interaction between the structure of the environment and the previously acquired knowledge of the learner. Similarly, \cite{haber2018emergence} created an agent with a world-model that learned to predict the consequences of the agent's actions, and a meta-cognitive self-model that tracks the performance of the world-model. The self-model was able to improve the world-model by adversarially challenging it. This caused the agent to autonomously explore novel interactions with the environment, leading to new behaviours and improved learning. In sum, there has already been some integration of adult and infant cognitive (neuro)science into active learning in ML \cite{gottlieb2018towards}.

This section has thus far focused primarily on attention allocation. However, other forms of active learning are worth mentioning. Particularly, the influence of obtaining a sense of agency plays an essential role in infants’ possibilities to actively shape their own learning environment. Knowledge about what one can do, allows intervention in the world in a manner optimised for acquiring new information about the environment \cite{lagnado2002learning, pearl2009causality}. As such, learning about one's body and agency has been studied both in infants \cite{bremner2016developing,rochat2000perceived, watanabe2011initial, zaadnoordijk2020movement} as well as in artificial systems \cite{diez2019sensorimotor, hafner2020prerequisites, zaadnoordijk2019match}. Although these topics might seem too embodied for traditional ML, the principles of active intervention for learning do not require a physical body and are thus of relevance nonetheless. Moreover, active exploration of objects has been shown to increase learning about those objects compared to passive observation \cite{ivaldi2013object, oakes2012manual}. The benefit of exploration and play is most visible in the flexibility and creativity of later metacognitive and problem-solving capacities \cite{caruso1993dimensions}.

\textbf{TL;DR:} Having opportunities to selectively influence their learning curriculum or interact with the environment in targeted ways is a crucial aspect to infant learning and has been proven successful when applied to artificial systems too. By taking previously encountered and encoded information into account, the mechanism can optimise for those inputs that maximally increase learning. Active learning has the potential to dramatically change the speed and quality of learning in DNNs, where it has been only sparsely incorporated until now.

\subsection{Babies learn from other agents}
\label{subsec:sociallearning}

In the previous section, we argued that infants do not learn passively. Here, we address the consequences of them not learning in a vacuum either. Infants have access to a rich input from other agents by observing them as well as by active engagement from them. %Active engagement of parents and other social agents in infants' lives has long been considered of great importance. 
Parental scaffolding, a process where parents are helping and guiding their infant to a greater or lesser extent depending on the infant's needs, has been shown to be uniquely important for infant development \cite{vygotsky1978mind,wood1976role}. For example, when infants play with a parent (or other more experienced person), infants' quality of play increases to an extent that could not be achieved by the infant alone \cite{vygotsky1978mind}. This heightened level of play is a consequence of the parent's attention and contingent reactions to the infant's actions \cite{bigelow2004role}. This means that scaffolding and guided play is both interactive as well as dynamic \cite{yu2018theoretical}, ensuring maximal support to the infant's learning process. During parent-child interaction, the parent can direct their attention to the infant's object of attention, or redirect the infant's attention to another object. Such joint attention episodes are considered `hot spots' for language learning \cite{tomasello1988role} and a `major contributor' to social development \cite{mundy2007attention}. Moreover, parental attention directing behaviours have been shown to influence infants' exploration behaviour \cite{belsky1980maternal} and cognitive development \cite{landry2000early, landry2006responsive}. While playing with infants, parents also tend to exaggerate their actions, which has obvious pedagogical benefits \cite{brand2002evidence, van2020motion}. This behaviour was coined `motionese' \cite{brand2002evidence} and is considered to be the action-counterpart to `motherese' \cite{newport1977motherese}. Motherese (or: infant-directed speech) is thought to help infants learn aspects of language (e.g. word segmentation and word recognition) due to its exaggerated use of changes in pitch and tempo \cite{singh2009influences, thiessen2005infant}. Thus, although parents do not generally explicitly teach their children in early infancy, they do tend to tailor their behaviour in such a way that allows the infant to better learn.

In addition to this active engagement from other agents, infants also learn from others by simply observing them. For example, \cite{provasi20019} found that infants who had observed an adult activate a knob were more successful at repeating this action than infants who only had had active experience with the knob themselves. Learning by observation and imitation has generally been widely explored and debated in the infant field \cite{meltzoff1997explaining, oostenbroek2018re, ray2011imitation} and has been shown to speed up learning of language acquisition \cite{kymissis1990history, poulson1991generalized}, action understanding and production \cite{elsner2007infants, hunnius2014you, paulus2014and}, social cognition \cite{meltzoff2003imitation}, and so on. Notably, some studies have shown that learning from observing others is modulated by infants' experience with how reliable or knowledgeable the other agent is \cite{poulin2011infants, stenberg2012infants, tummeltshammer2014infants}.

Inspired by human learning, scaffolding and other forms of human pedagogy have been successfully adopted in (developmental) robotics \cite{nagai2009computational, otero2008teaching, rohlfing2006can, ugur2011learning}. However, one might question whether there is a place for these approaches in unsupervised ML. Would parenting an algorithm not make it supervised learning? We propose that it is something in between. Although parents at times indeed provide direct feedback to infants, their importance for learning is much rather driven by their broader perspective. By being a sort of meta-learner, they can provide infants with the push they need to enter a new, steeper learning curve (see e.g. \cite{provasi20019}). While there is some work on teaching in artificial intelligence research \cite{bieger2017pedagogical}, little attention has been directed to using an `expert' network to guide (but not directly instruct) a `novice' network. Training unsupervised ML networks first on auxiliary tasks to boost performance on a target task (see e.g. \cite{ruder2017overview}) and curriculum learning (see Sec.~\ref{subsec:structuredinput}), may be ML's closest procedures to parental scaffolding as these can be re-conceptualised as the researcher or engineer scaffolding the network. 

\textbf{TL;DR:} Other agents in infants' everyday life provide essential scaffolds to their learning process. In unsupervised ML, parental scaffolding has not been widely adopted. However, the successes gained by training networks first on auxiliary tasks before moving onto the target task suggests that this is a space worth exploring.
As concluded by \cite{bonawitz2016computational} on computationally modelling infant learning: ``\textit{[...] to understand power of children’s learning, it is important to investigate it in a social context.}'' (p. 98)

\section{Discussion}
\label{sec:discussion}

Unsupervised ML approaches benefit from flexibly and efficiently learning flexible and efficient representations. Here we have argued that developmental science offers a unique source of inspiration; infants learn effective, generalised and transferable representations from relatively small numbers of quite heterogeneous examples. We have presented five core insights from developmental science that we believe can make a fundamental difference to representation learning in ML. We have focused especially on infant learning in the first year of life as their learning process also requires them to learn useful representations from unlabelled data. Throughout the paper, we have spelled out to what extent components of infant learning are already mirrored in ML algorithms, and where further steps can reasonably be made. Improving the quality, flexibility and efficiency of learned representations will directly translate to improved ML performance. 

We demonstrated that although infants and ML systems share commonalities (e.g., both perform statistical learning), the currently prevailing practices in ML---to remove targeted interference in the learning process as much as possible and leave everything to be learned to the data itself \cite{bengio2013representation}---stands in stark contrast to infant learning. Infants' input has been found to be optimised for learning about specific features of the input. When comparing the five lessons on infant learning to current approaches in ML, two overarching insights can be extracted: 
\begin{enumerate}
\item 
There is more structure to constrain and guide infants' learning processes (Sec.~\ref{subsec:architecture} and \ref{subsec:structuredinput}).
\item Infants' learning opportunities are more flexible (Sec.~\ref{subsec:activelearning}) and richer (Sec.~\ref{subsec:multiplestreams} and \ref{subsec:sociallearning}).
\end{enumerate}
Factors like innate biases, saliency, curiosity, and development over time all play an important role in shaping infants' learning curriculum and contribute to the speed and flexibility with which infants learn. Reflecting these insights back into ML, they cast significant doubt on the assumption that `the data will fix it' is indeed the most efficient and effective approach to training neural networks. 

Neither the developmental nor the ML research presented in this paper is exhaustive. Nonetheless, we hope to have provided a representative sample of prior work on the topics we have addressed. Importantly, the five lessons presented here were chosen based on their potential to qualitatively improve the next generation of unsupervised ML algorithms---either by introducing previously unconsidered aspects, or by reinforcing and helping to further evolve ongoing work---as well as on their integration with current ML implementations. By focusing on this intersection, we aim to increase the likelihood that these lessons can be meaningfully considered in theory and implementations.

The argument to take inspiration from human (infant) learning that we and others have made rests on the observation that human learning leads to robust representations that can be flexibly used in various tasks with an acceptable to excellent level of performance across the board. Clearly, some scepticism is warranted as to whether adopting insights from infant learning will be equally valuable for all ML purposes. It is possible that some of the lessons provide an advantage across domains whereas others might turn out to be particularly beneficial for specific tasks. Taking this paper as a conceptual anchor, future research will explore the exact interactions of each of the given insights from infant learning with its counterpart(s) in ML.

\section*{Acknowledgements}
This work was supported by the ERC Advanced Grant FOUNDCOG, No. \#787981, awarded to Rhodri Cusack and the MSCA Individual Fellowship InterPlay, No. \#891535, awarded to Lorijn Zaadnoordijk.

\bibliographystyle{plain}
\bibliography{LessonsDevSci} 

\end{document}